\documentclass[10pt,twocolumn,letterpaper]{article}

\usepackage{cvpr}
\usepackage{times}
\usepackage{epsfig}
\usepackage{graphicx}
\usepackage{amsmath}
\usepackage{amssymb}
\usepackage[ruled]{algorithm2e}
\usepackage{comment}
\usepackage[british,UKenglish,USenglish,english,american]{babel}
\usepackage{authblk}

\usepackage[pagebackref=true,breaklinks=true,letterpaper=true,colorlinks,bookmarks=false]{hyperref}

\newcommand{\Sref}[1]{Section~\ref{#1}}

\newcommand{\eref}[1]{Eq.~(\ref{#1})}
\newcommand{\fref}[1]{Fig.~\ref{#1}}

\graphicspath{ {images/} }

\cvprfinalcopy 

\def\cvprPaperID{****} 

\begin{document}

\title{Arbitrary Style Transfer with Deep Feature Reshuffle}

\author{Shuyang Gu$^{1}$\thanks{Equal contribution. This work is done when Shuyang Gu and Congliang Chen are interns at Microsoft Research Asia.} \qquad Congliang Chen$^{2}$$^{*}$ \qquad Jing Liao$^{3}$ \qquad Lu Yuan$^{3}$ \vspace{1pt}\\
$^{1}$University of Science and Technology of China \qquad $^{2}$Peking University \qquad $^{3}$Microsoft Research\qquad\qquad\\
\hspace{0.1in}{\tt\small gsy777@mail.ustc.edu.cn} \qquad  {\tt\small chcoliang@pku.edu.cn} \qquad  {\tt\small \{jliao,luyuan\}@microsoft.com} \\
}

\maketitle

\begin{abstract}

This paper introduces a novel method by reshuffling deep features (\ie, permuting the spacial locations of a feature map) of the style image for arbitrary style transfer. We theoretically prove that our new style loss based on reshuffle connects both global and local style losses respectively used by most parametric and non-parametric neural style transfer methods. This simple idea can effectively address the challenging issues in existing style transfer methods. On one hand, it can avoid distortions in local style patterns, and allow semantic-level transfer, compared with neural parametric methods. On the other hand, it can preserve globally similar appearance to the style image, and avoid wash-out artifacts, compared with neural non-parametric methods. Based on the proposed loss, we also present a progressive feature-domain optimization approach. The experiments show that our method is widely applicable to various styles, and produces better quality than existing methods. 
\end{abstract}

\section{Introduction}
This process of rendering a content image in the style of another image is referred to as Style Transfer. The problem of style transfer has its origin from non-photo-realistic rendering~\cite{kyprianidis2013state}, and is closely related to texture synthesis and transfer~\cite{efros2001image,efros1999texture,elad2016style}. These methods typically rely on low-level statistics and often fail to capture semantic structures. Recently, the work of Gatys et al.~\cite{gatys2015neural} opened up a new field called Neural Style Transfer, which uses Convolutional Neural Network (CNN)~\cite{krizhevsky2012imagenet} to change the style of an image while preserving its content. It is flexible enough to combine content and style of arbitrary images.

Style transfer is receiving increasing attention from computer vision researchers because it involves two interesting topics: image representation and image synthesis. Some early representations, like multi-resolution~\cite{de1997multiresolution}, pyramid~\cite{heeger1995pyramid}, wavelet~\cite{portilla2000parametric}, used in traditional texture synthesis and transfer, are mainly for statistics matching. The recent work~\cite{gatys2015neural} showed that the representations of image content and style were separable by variant CNN convolutional layers. Moreover, the representation provides the possibility for image decoupling and recombining.

Image synthesis methods, whether traditional or neural, can be broadly categorized as \emph{parametric} and \emph{non-parametric}. Specifically, for neural methods, the parametric methods match the global statistics of deep features, like Gram matrix ~\cite{gatys2015neural,johnson2016perceptual} and its approximates~\cite{li2017universal,lu2017decoder}, mean and variance~\cite{huang2017arbitrary,dumoulin2016learned}, histogram \cite{wilmot2017stable}; while the non-parametric methods~\cite{chuanli2016,li2016precomputed,chen2016fast,liao2017visual} directly find neural patches similar to the given example. However, to the best of our knowledge, there are no work connecting these neural methods to form a complementary solution.

\begin{figure}[t]
\centering
\footnotesize
\setlength{\tabcolsep}{0.003\linewidth}
\begin{tabular}{cccc}
 \includegraphics[width=0.22\linewidth]{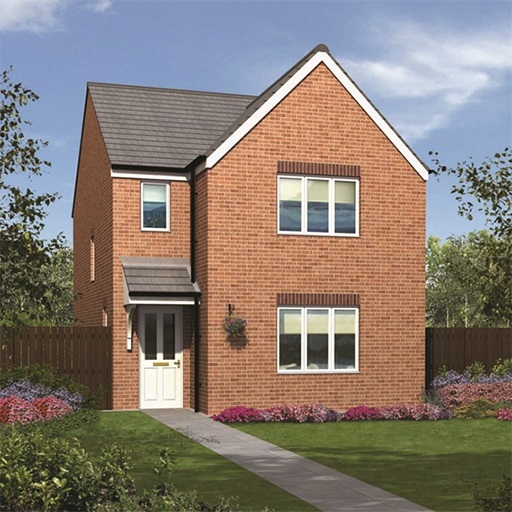} &  \includegraphics[width=0.22\linewidth]{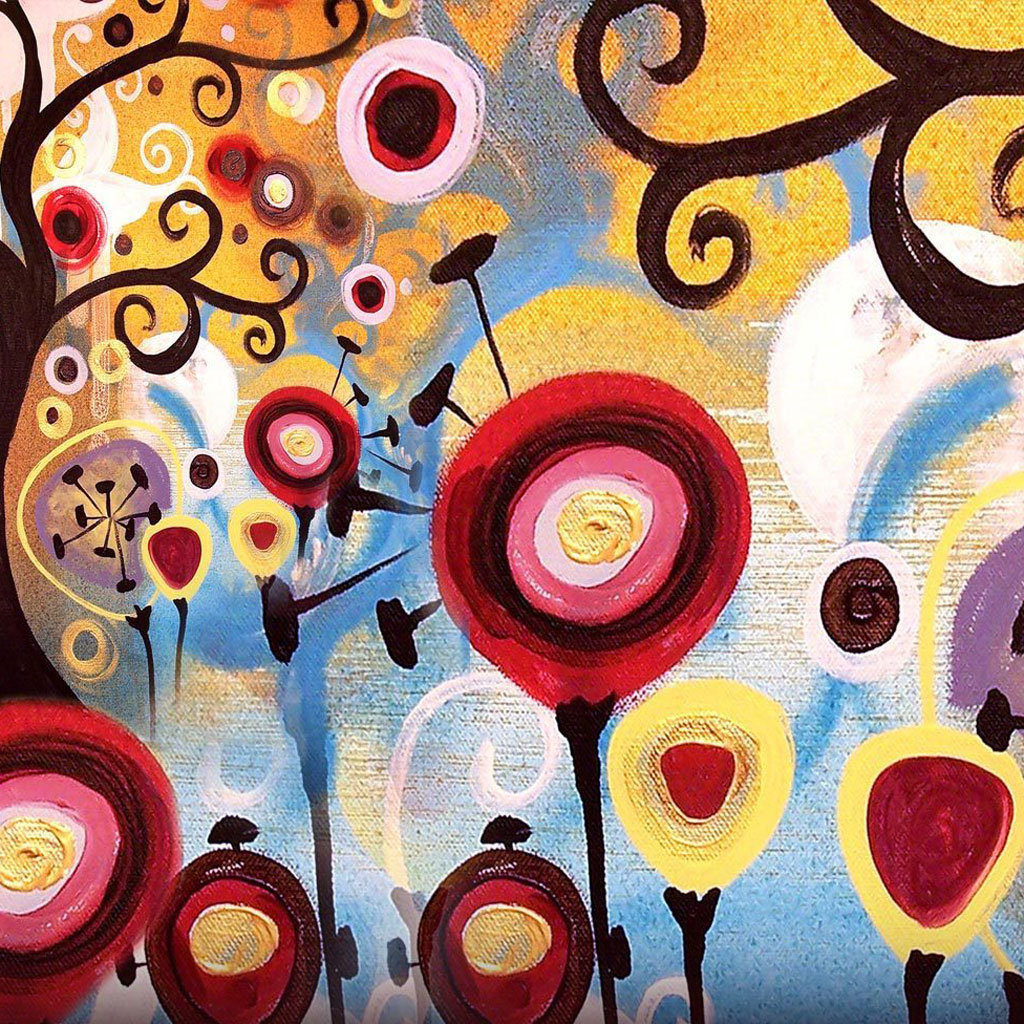} & \includegraphics[width=0.22\linewidth]{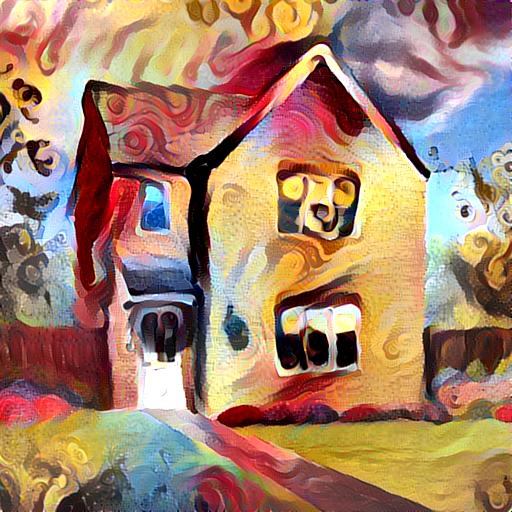} & \includegraphics[width=0.22\linewidth]{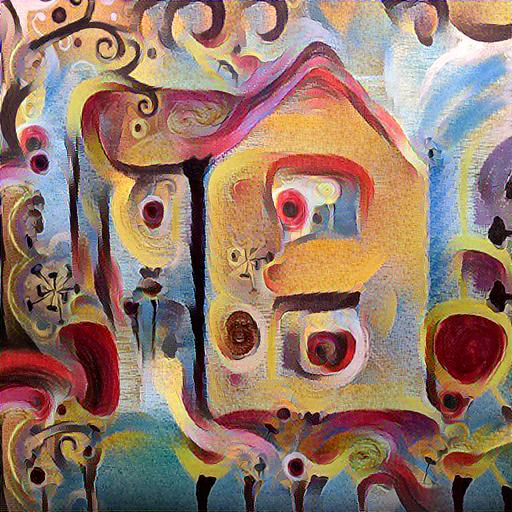}\\
  Content & Style & Parametric\cite{gatys2015neural} &Non-parametric\cite{chuanli2016}\\
\includegraphics[width=0.22\linewidth]{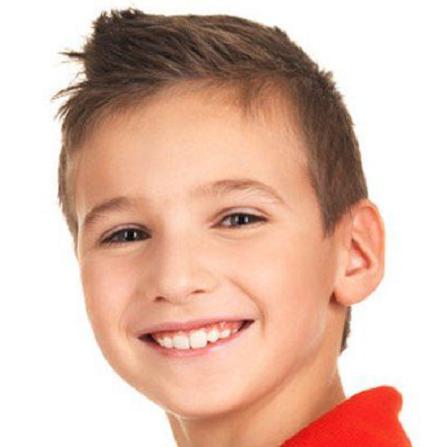} &  \includegraphics[width=0.22\linewidth]{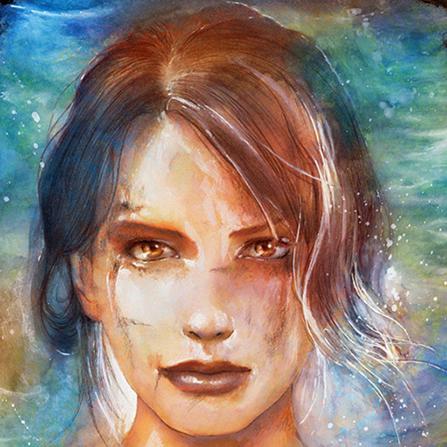} & \includegraphics[width=0.22\linewidth]{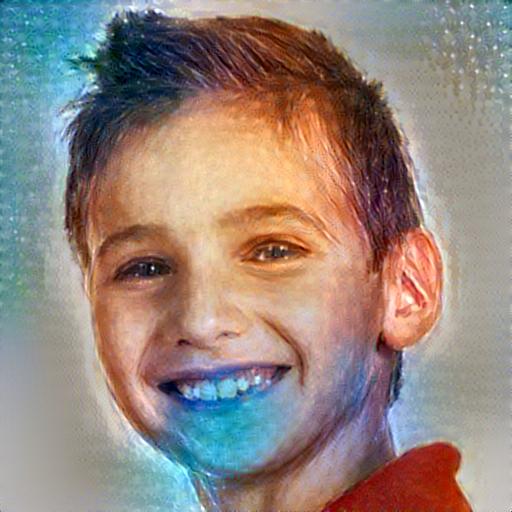} & \includegraphics[width=0.22\linewidth]{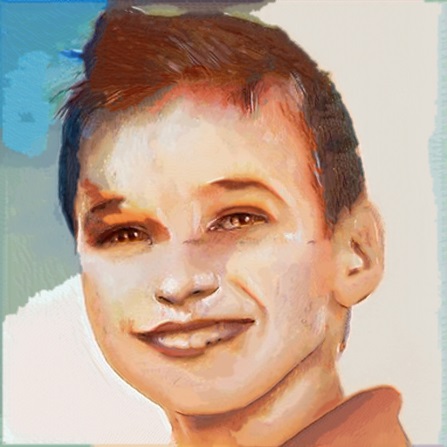}\\

Content & Style & Parametric\cite{gatys2015neural} & Non-parametric\cite{liao2017visual} \\
\end{tabular}
   \caption{Parametric style transfer fails to preserve some local texture patterns of the style, e.g. circles in the upper row, and it have weak spatial constraint, like rendering background colors in the face region (lower row).  }\vspace{-1.3em}
\label{fig:introduction_parametric}
\end{figure}

\begin{figure}[t]
\centering
\footnotesize
\setlength{\tabcolsep}{0.005\linewidth}
\begin{tabular}{cccc}
 \includegraphics[width=0.22\linewidth]{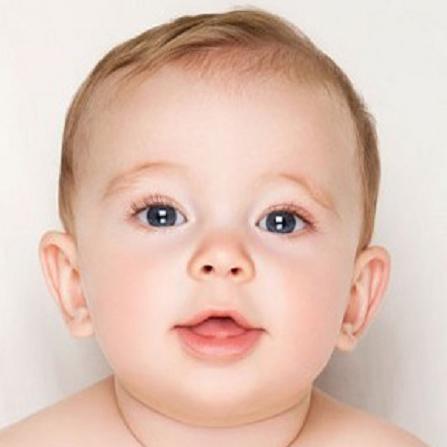} &  \includegraphics[width=0.22\linewidth]{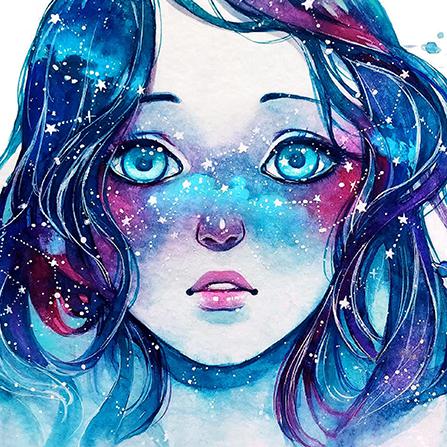} & \includegraphics[width=0.22\linewidth]{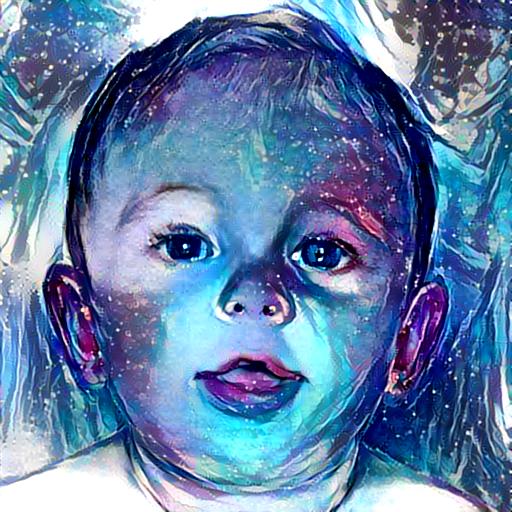} & \includegraphics[width=0.22\linewidth]{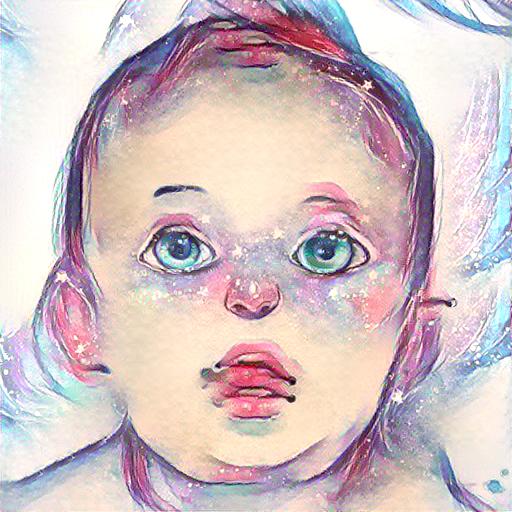}\\
  Content & Style  & Parametric\cite{gatys2015neural} & Non-parametric\cite{chuanli2016}  \\
\includegraphics[width=0.22\linewidth]{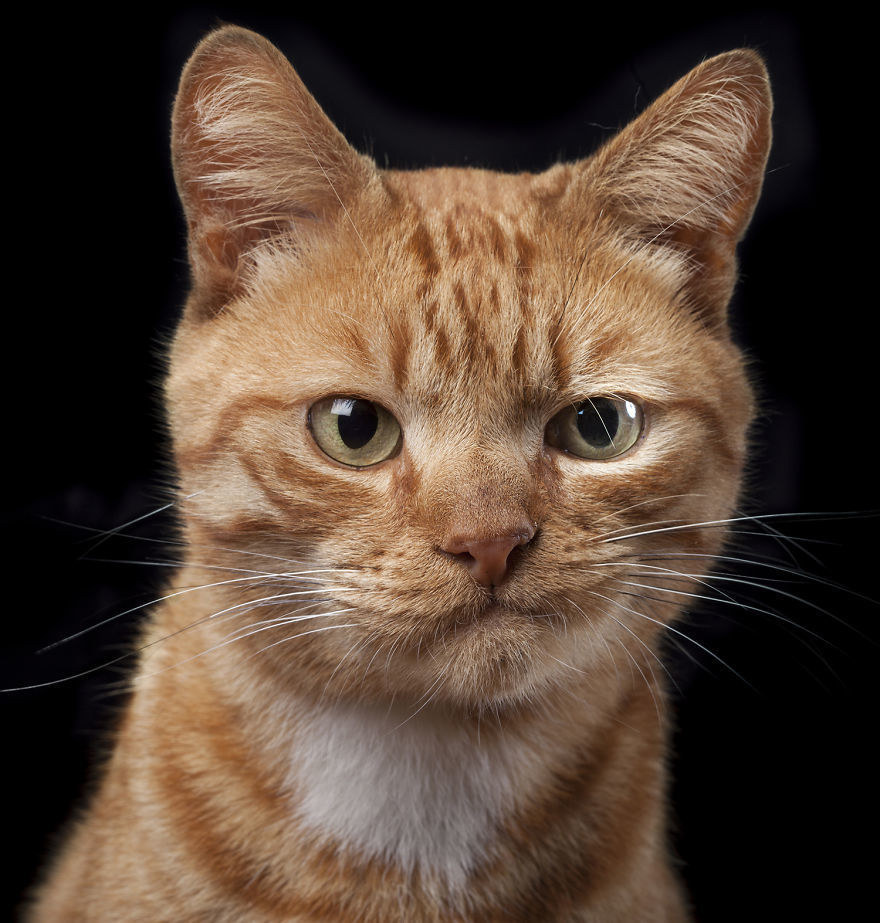} &  \includegraphics[width=0.165\linewidth]{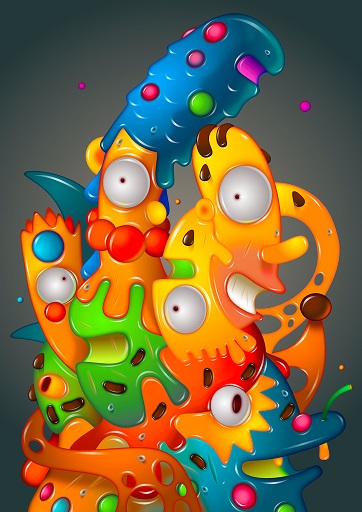} & \includegraphics[width=0.22\linewidth]{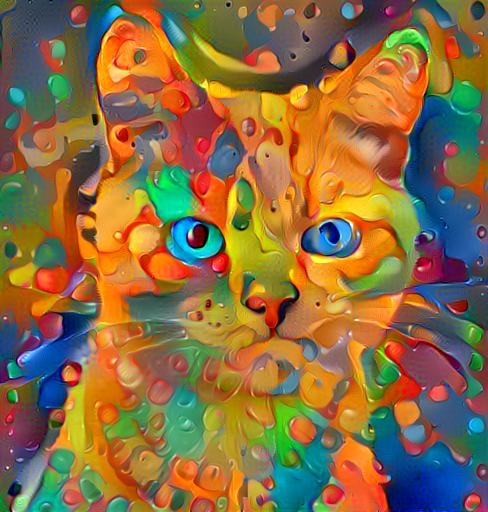} & \includegraphics[width=0.22\linewidth]{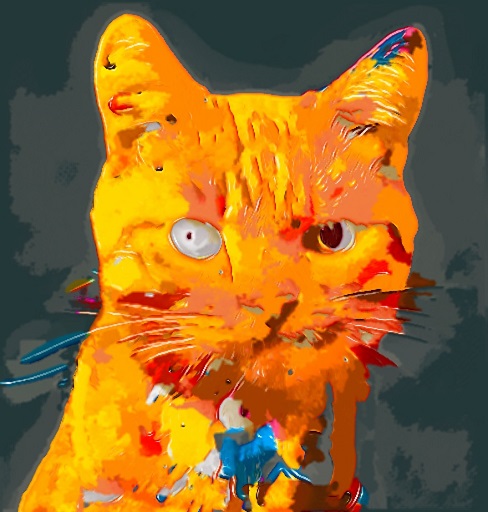}\\
Content & Style & Parametric\cite{gatys2015neural}& Non-parametric\cite{liao2017visual}  \\
\end{tabular}
    \caption{Non-parametric style transfer is sometimes globally less similar to the style (upper row) and repetitively uses the same patches to cause wash-out artifacts (lower row). }\vspace{-1.7em}
\label{fig:introduction_non_parametric}
\end{figure}

The \emph{neural parametric} models (\eg,~\cite{gatys2015neural}) yields results, preserving the content of image and the overall looking of the artwork. However, the models will distort local style patterns (shown in the first row of~\fref{fig:introduction_parametric}) or cannot obtain locally semantic-level transfer (\eg, eye-to-eye in the second row of~\fref{fig:introduction_parametric}). The \emph{neural non-parametric} models (\eg,~\cite{chuanli2016}) can address these issues well, but their example matching uses a greedy optimization, causing the decreasing in the richness of the style patterns (shown in the first row of~\fref{fig:introduction_non_parametric}), and introducing wash-out artifacts \cite{jamrivska2015lazyfluids} (see the second row of~\fref{fig:introduction_non_parametric}). It suggests that such \emph{neural non-parametric} models should consider global constraint, borrowing from \emph{neural parametric models}.

In this paper, we propose a novel neural style transfer algorithm which owns the advantages of both neural parametric and non-parametric methods. This is achieved by \emph{deep feature reshuffle}, which refers to spatially rearranging the position of neural activations. We reshuffle the features of the style image according to the content image for style transfer. On one hand, the feature reshuffle enforces the distribution of style patterns, between the transferring result and the style image, to be globally consistent. It can be theoretically proved that reshuffling features of style image equals to the optimization of Gram matrices, a commonly used statistics in \emph{neural parametric} models. On the other hand, a certain type of reshuffling features can help achieve locally semantic matching between images, as well as \emph{neural non-parametric} models.

We reformulate the objective function of neural style transfer in the fashion of reshuffle, and then connects both kinds of methods. To avoid exhaustively optimizing the energy function in image domain as similar as~\cite{gatys2015neural}, we propose a novel optimization method in feature domain. We can progressively recover features from high-level layers to low-level ones, and train a decoder to convert recovered features back to the image. This way is more efficient.

Our experiments show that this method can effectively accomplish the transfer for arbitrary styles, yield results with global similarity to the style and local plausibility. We summarize main contributions as follows: \vspace{-0.4em}

\begin{itemize}
\item We provide a new understanding of \emph{neural parametric} models and \emph{neural non-parametric} models. Both can be integrated by the idea of \emph{deep feature reshuffle}. \vspace{-0.2em}
\item We define a new energy function based on \emph{deep feature reshuffle}, which is simple, flexible, and better than either \emph{neural parametric} or \emph{non-parametric} methods. \vspace{-0.2em}
\item We train a new level-wise decoder to allow us efficiently optimize our feature-domain energy function in a pyramid manner.
\end{itemize}

\section{Related Work}

The problem of style transfer involves two sub-problems: representation and synthesis. Inspired by the success of CNN in style transfer, we also use neural representation for image decoupling, and better matching. In this paper, we focus on synthesis problem, which can be categorized as parametric and non-parametric.

In fact, parametric and non-parameter synthesis methods early occur at texture synthesis and transfer. Parametric methods \cite{de1997multiresolution,heeger1995pyramid,portilla2000parametric}
start from random noise, and then iteratively update it until the desired global statistics is satisfied. However, it is challenging to find a proper statistical model for representation and fine matching. By contrast, non-parametric models \cite{efros2001image,efros1999texture,wei2000fast} use a simple patch representation (\eg, color~\cite{efros2001image,hertzmann2001image,ashikhmin2003fast}, curvilinear features \cite{wu2004feature}, edge~\cite{kaspar2015self} and its orientation \cite{lee2010directional}), and find the most similar patches by nearest neighbor search. All above methods only use the low-level features for synthesis, which limits to capture semantic structures.

Gatys et al.~\cite{gatys2015neural} pioneer the neural texture synthesis and style transfer by successfully applying CNN (pre-trained VGG networks \cite{simonyan2014very}) to this problem. The spirit of their synthesis method is parametric, which statistically matches both the content and style features by their Gram matrices. The solution is general to varies of artistic styles, and begins to considering semantic structures. To improve the quality, some complimentary information is incorporated into the statistical model, including spatial correlation~\cite{sendik2017deep}, face guidance~\cite{selim2016painting}, user controls~\cite{champandard2016semantic,gatys2016controlling}, and segmentation masks~\cite{luan2017deep}. To accelerate, a feed-forward generative network~\cite{johnson2016perceptual,ulyanov2016texture,dumoulin2016learned,dong2017stylebank} is directly learnt instead, but they are still limited to a fixed number of pre-trained styles. More recently, some work~\cite{li2017universal,huang2017arbitrary,lu2017decoder} further allow arbitrary style transfer in feedforward networks. The backend idea is to match the statistics of content features at intermediate layers to that of the style features, and then train a decoder to turn features to the image. And \cite{chen2018stereoscopic,chen2017coherent} further extend this kind of method to video and stereoscopic 3D style transfer.

Non-parametric neural style transfer method is firstly proposed by Li et al.~\cite{chuanli2016}.
They reformulate the style transfer based on Markov Random Field (MRF): searching local neural patches from the style image to satisfy the local structure prior of content image. Compared with neural parametric methods, this method can reproduce local textures more faithfully. They also train a Generative Adversarial Networks (GAN) to accelerate the optimization \cite{li2016precomputed}. Chen et al. \cite{chen2016fast} use de-VGG networks and the patch-based method for fast arbitrary style transfer. Another representative work is called Deep Analogy \cite{liao2017visual}, which proposes accurate semantic-level patch match algorithm by considering bidirectional constraint and pyramids refinement. Our method is based on non-parametric model, sharing all the advantages, like local similarity to style and semantic-level transfer. Moreover, it allows some global constraints to avoid washout artifacts, and obtains global consistency, as similar as parametric methods.

\section{Understanding Neural Style Transfer}

In this section, we explore the relationship between neural parametric method (\eg,~\cite{gatys2015neural}) and neural non-parametric method (\eg,~\cite{chuanli2016}). Then, we realize that the \emph{feature reshuffle} can theoretically be a complementary solution for both methods.

For the task of style transfer, we want to generate a stylization result $I_o$, given the content image $I_c$ and the style image $I_s$. For simplicity, we suppose $I_o$, $I_c$ and $I_s$ are with the same size\footnote{For different size, the principle still holds true, when these features and terms are normalized according to feature size.}, and consider the single layer feature of these image, which are denoted as $F_o$, $F_c$ and $F_s$ respectively. $F\in \Omega^{c\times h\times w}$ is indeed 3D tensor, where $c,h,w$ denote channel number, height and width respectively.

\paragraph{Neural Parametric.} In \cite{gatys2015neural}, the energy function consists of a content term $\mathcal{L}_{cont}$ and a style term $\mathcal{L}_{sty}$ \footnote{We ignore the image regularization term here, because it is common, and only takes effects to subtle noise. Its effect to our analysis can be negligible.}:
\begin{equation}
\mathcal{L}_{total}=\alpha \mathcal{L}_{cont} + (1-\alpha) \mathcal{L}_{sty},
\label{eq:loss}
\end{equation}
where $\alpha$ is the tradeoff to balance content and style.

The content loss $\mathcal{L}_{cont}$ is defined by the feature difference between the content image $I_c$ and the (yet unknown) stylized image $I_o$:
\begin{equation}
\mathcal{L}_{cont}=||F_o-F_c||^2_{\rm{F}},
\label{eq:content}
\end{equation}
where $||\cdot||^2_{\rm{F}}$ denotes Frobenius norm. 

For the style loss $\mathcal{L}_{sty}$, Gatys et al.~\cite{gatys2015neural} uses Gram matrix $G(i,j)$ to obtain correlations between filter responses. It was used to measure texture correlation in texture synthesis algorithm~\cite{gatys2015texture}. Gram matrix $G(i,j)$ is defined as the inner product between the $i\_\text{th}$ and $j\_\text{th}$ feature channels:
\begin{equation}
G(i,j)=\sum\mathop{}_{p}F(i,p)F(j,p),
\label{eq:gram}
\end{equation}
where $p$ denotes the 2D spatial location in feature map and $G\in \Omega^{C\times C}$. Then, the style loss $\mathcal{L}_{sty}$ (also called global style loss later) is defined by the difference between Gram matrices of $F_o$ and $F_s$:
\begin{equation}
\mathcal{L}_{sty}=||G_o-G_s||^2_{\rm{F}}\footnote{It should be $\mathcal{L}_{sty}=\frac{1}{\mathcal{Z}}||G_o-G_s||^2_{\rm{F}}$, where $\mathcal{Z} = 4\times c^2 \times h^2 \times w^2$. We leave out $\mathcal{Z}$ for simplicity, and it won't effect our analysis.}
\label{eq:global}
\end{equation}

\paragraph{Neural Non-parametric.} In~\cite{chuanli2016}, the energy function is also defined with two terms:
\begin{equation}
\mathcal{L}_{total}=\alpha \mathcal{L}_{cont} + (1-\alpha) \mathcal{L}_{match}.
\label{eq:loss}
\end{equation}

The content loss $\mathcal{L}_{cont}$ is identical to~\eref{eq:content}; while the style loss $\mathcal{L}_{match}$ measures the neural patch-based similarity. Let $\Psi(F)$ denote the list of all local patches extracted from feature map $F$. Each \emph{neural patch} centered at the location $p$ of feature $F$ is indexed as $\Psi_p(F)$. The loss $\mathcal{L}_{match}$ (also called local style loss later) is defined as:
\begin{equation}
\mathcal{L}_{match}=\sum\mathop{}_p||\Psi_{p}(F_o)-\Psi_{\text{NN}(p)}(F_s)||^2_{\rm{F}},
\label{eq:local}
\end{equation}
where $\text{NN}(p)$ is the index of the patch in $\Psi(F_s)$ which is the most similar to $\Psi_{p}(F_o)$. The best matching index is calculated using normalized cross-correlation over all local patches in the style feature $F_s$:
\begin{equation}
	\text{NN}(p)=\mathop{\arg{\max}}_{p'=1,2,\dots,\Theta} \frac{\Psi_{p}(F_o) \cdot \Psi_{p'}(F_s)}{||\Psi_{p}(F_o)||^2_{\rm{F}} \cdot ||\Psi_{p'}(F_s)||^2_{\rm{F}}},
	\label{eq:nn}
\end{equation}
where the operator $\cdot$ denotes inner product, and $\Theta = h\times w$.\\

\noindent Both methods share the same content loss, but have different style loss terms. The global style loss (\eref{eq:global}) measures global statistics, but ignores the spatial layout of features~\cite{li2017demystify}. On the contrast, the local style loss (\eref{eq:local}) encourages to find optimal feature layout for each local patch individually (\eref{eq:nn}), but without global constraint.

\paragraph{Neural Feature Reshuffle.}

%



Can an optimal feature $F_o$ be achieved to satisfy both global and local style terms simultaneously? Yes, we find that \emph{feature reshuffle} can theoretically be an ideal condition, which can make both global and local style terms be zero.

Specifically, \emph{feature reshuffle} means that we permute the spatial location of feature map $F_s$ to reconstruct a new feature map $F_o$. Let $\text{SF}(p)$ be the permuted location of the feature corresponding to the original location $p$. By feature reshuffle, the reconstructed feature map $F_o$ is denoted as:
\begin{equation}
F_o(p)=F_s(\text{SF}(p)), \;\text{ and }\; F_s(p)=F_o(\widetilde{\text{SF}}(p)),	
\label{eq:resh}
\end{equation}
where $\widetilde{\text{SF}}(p)$ is the inverse reshuffle function. In other words, $\text{SF}(p)$ is a bijection: for each location in $F_o$ there exists exactly one location in $F_s$ corresponding to it.

We can deduce the global style loss (in \eref{eq:global}) with the reshuffle solution (see in \eref{eq:resh}) as:
\begin{displaymath}
\begin{split}
&\mathcal{L}_{sty}= \sum_{i,j}||\sum_{p}F_o(i,p)F_o(j,p)-\sum_{p}F_s(i,p)F_s(j,p)||^2\\
&=\sum_{i,j}||\sum_{p}F_s(i,\text{SF}(p))F_s(j,\text{SF}(p))-\sum_{p}F_s(i,p)F_s(j,p)||^2\\
&=0,
\end{split}
\label{eq:global1}
\end{displaymath}
which indicates reshuffling features of style does not effect Gram matrices of style features, leading the global style loss to be zero.

According to \eref{eq:resh}, the feature $F_o(p)$ at each point $p$ is reconstructed by $F_s(SF(p))$. When the patch size is $1\times 1$, $\Psi_{p}(F_o)$ can be denoted as $F_o(p)$. Then, the nearest neighbor of $F_s(SF(p))$ to be found in the feature map $F_s$ is just itself, \ie, $F_s(SF(p)) = F_s(NN(p)) = \Psi_{NN(p)}(F_s)$. Based on these deductions, we can rewrite \eref{eq:local} as

\begin{displaymath}
\mathcal{L}_{match}=\sum_{i}||\sum_{p}(F_o(i,p)-F_s(i,\text{SF}(p)))||^2 = 0.
\label{eq:global1}
\end{displaymath}\vspace{-1.0em}

%
%

As summary, the above theoretical derivation demonstrates feature reshuffled from the style image simultaneously minimizes both global style loss and local style loss (when patch size is $1\times1$).

\section{Method}

Based on the idea of \emph{deep feature reshuffle}, we propose a novel neural style transfer algorithm, which integrates global and local style losses in the whole objective function. We first present a new style loss called \emph{reshuffle loss}, which would be combined with content loss as well. Then, we show the optimization in a single feature layer. Such an optimization can be done in image domain, similar to the manner of~\cite{gatys2015neural}, by iteratively forward and backward passing the networks. For acceleration, we propose two efficient ways: 1) the optimization can be done in feature domain, and does not need to back propagate to the image at every time; 2) we progressively optimize the features across multiple layers, and the exhaustive patch match in the fine layer can be guided by the matching result in the coarse layer.

\subsection{Reshuffle Loss Function}

\begin{figure}[t]
\centering
 \includegraphics[width=0.95\linewidth]{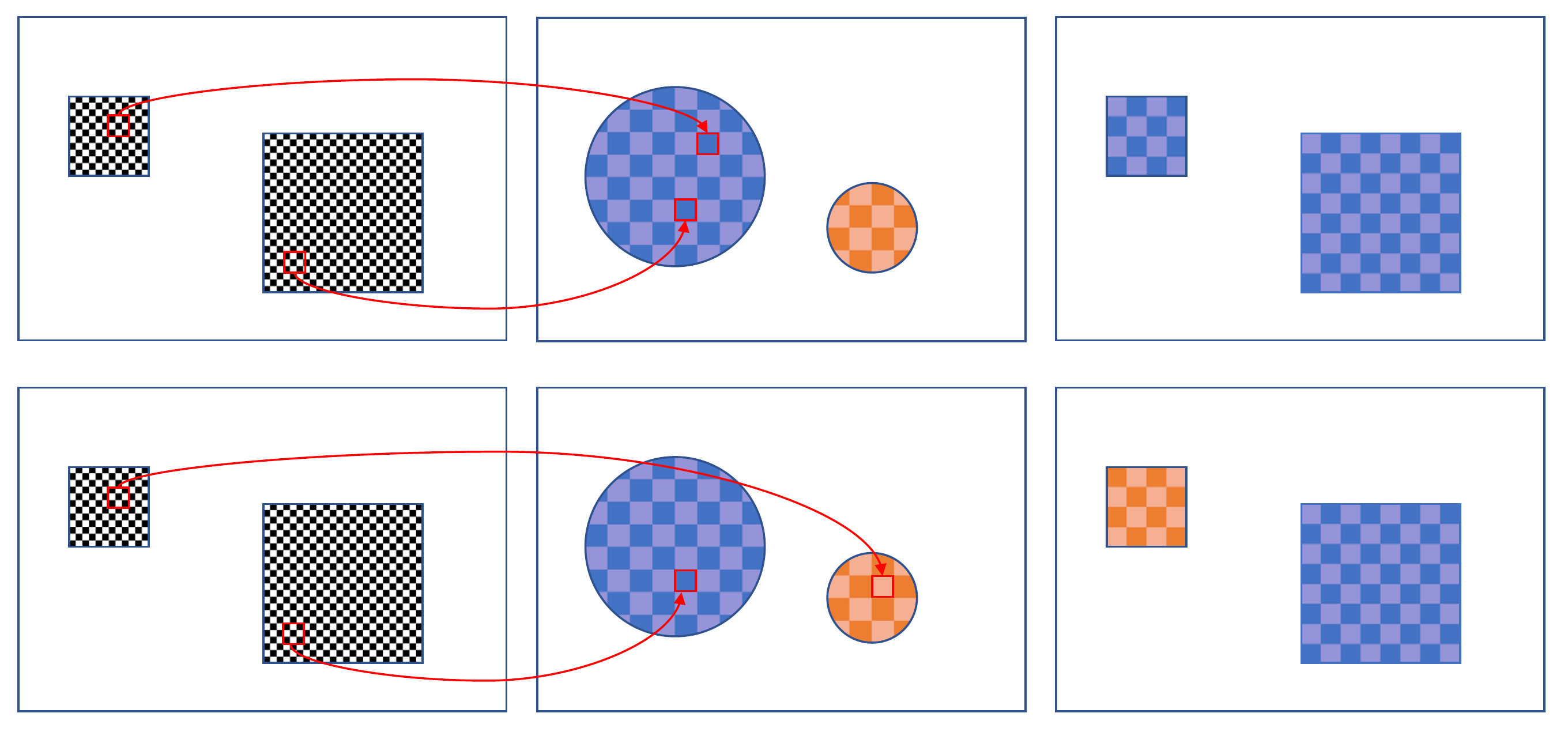}
 \footnotesize
\begin{tabular}{ccc}
 \quad  \quad Target feature & \quad  \quad  Source feature &  \quad  Reconstructed feature
\end{tabular}
    \caption{The comparison between normal NN search (upper row) and our NN search constrained by reshuffle (lower row). Ours strictly requires each patch in the source to be only mapped once. Notice that the bottom only shows a possible solution of reshuffle.}
\label{fig:reshuffle}\vspace{-0.7em}
\end{figure}

We define a new \emph{reshuffle loss} for the style loss, which only slightly modifies the local style loss term (see \eref{eq:local}):
\begin{equation}
\mathcal{L}_{shuf}=\sum\mathop{}_{p}||\Psi_{p}(F_o)-\Psi_{\text{NNC}(p)}(F_s)||^2_{\rm{F}},
\label{eq:shuf}
\end{equation}
where the original nearest neighbor (NN) search $\text{NN}(p)$ is replaced by a new function $\text{NNC}(p)$. It is also the NN search, but constrained by the times of patch usage. For strict reshuffle, we require each patch in the source to be only mapped once, as shown in \fref{fig:reshuffle}.

Indeed, sometimes such one-usage constrain is too harsh. For example, the content image has two faces but the style image only has one. We relax the constraint in our energy term to allow more times of patch usage. Although it will sacrifice the global term to some extent, it greatly improve the robustness. A hard cutoff of usage is difficult to be found for every case. Instead, our constrained NN search function softly encourages the uniform usage of patch, inspired by \cite{kaspar2015self,fivser2016stylit}, which considered similar constraints in image domain, and is defined as:
\begin{equation}
\begin{split}
\text{NNC}(p)=\mathop{\arg\max}_{p'=1,2,\dots,\Theta} \ \ (\frac{\Psi_{p}(F_o) \cdot \Psi_{p'}(F_s)}{||\Psi_{p}(F_o)||^2_{\rm{F}} \cdot ||\Psi_{p'}(F_s)||^2_{\rm{F}}}\\
-\lambda\frac{\Gamma(\Psi_{p'}(F_s))}{R\times R}),\;\;
\label{eq:nnc}
\end{split}
\end{equation}
where $R$ is the patch size. $\Gamma(p)$ keeps track how many times of each pixel has been used in all patches covering it, and $\Gamma(\Psi_{p})$ refers to the total usages of a patch normalized by its area $R\times R$, namely,
\begin{equation}
\Gamma(\Psi_{p})= \sum\mathop{}_{p\in \Psi_{p}} \frac{\Gamma(p)}{R\times R}.
\label{eq:usage}
\end{equation}
This term requires each pixel to be used only once, as possible as it can. $\lambda$ controls the relative contribution of uniformity enforcement. The $\text{NNC}(p)$ can be optimized with the EM-like algorithm described in \cite{kaspar2015self,fivser2016stylit}, which extends PatchMatch algorithm \cite{barnes2009patchmatch} to keep track of the usage as well.

\subsection{Single Layer Optimization}
\label{sec:sig}
The objective function for single layer is defined as:
\begin{equation}
\mathcal{L}_{total}=\alpha\mathcal{L}_{cont}+ (1-\alpha) \mathcal{L}_{shuf}.
\label{eq:single}
\end{equation}
The most direct solution is to optimize it in image domain. Similar to \cite{gatys2015neural,chuanli2016}, we suppose the output image $I_o$ to be either the content image or random noise initially. Then we pass it to the VGG19 network (pre-trained on ImageNet for object recognition), and get the feature maps $F_o^l$ at $relul\_1$ layer. We optimize $\text{NNC}$ with the feature map $F_o^l$ and $F_s^l$, shown in \eref{eq:nnc}. The gradient of the energy function $\mathcal{L}_{total}$ (in~\eref{eq:single}) is then computed and back propagated to update $I_o$. Such processing always needs hundreds of iterations to converge by gradient decent method (\eg, L-BFGS \cite{zhu1997algorithm}). Hundreds times of foward-backward passes of networks and NN search make it prohibitively slow.

An alternatively fast solution is to directly optimize $F_o^l$ in feature domain, and then reverse it to the output image by training VGG-like image decoder. Here, we first discuss how to get $F_o^l$ by minimizing \eref{eq:single}. Considering \eref{eq:nnc} is not directly differentiable, the objective function (\eref{eq:single}) is hard to be optimized by a standard solver like SGD. To solve this issue, we adopt an iterative EM-like algorithm, which was used in~\cite{kaspar2015self} with good convergency. Specifically, $F_o^l$ is initialized to the content features, $F_o^{l,(0)}=F_c^l$. In the E-step of each iteration $i$, the constrained NN field $\text{NNC}^{l,(i)}$ is computed by matching $F_o^{l,(i-1)}$ and $F_s^l$ (see~\eref{eq:nnc}). We then get the feature map $F_s^l(\text{NNC}^{l,(i)})$, by warping $F_s^l$ with $\text{NNC}^{l,(i)}$ and average-voting overlapping neighbor patches at each location. In the M-step, we obtain the blended feature map $F_o^{l,(i)}$ by a linear combination: $F_o^{l,(i)}=\alpha F_c^l+(1-\alpha)F_s^l(\text{NNC}^{l,(i)})$, according to \eref{eq:single}. After several iterations (less than $10$ normally), $F_o^{l,(i)}$ will converge to the optimal feature $F_o^l$.

Once $F_o^l$ is achieved, we will get the output image $I_o$ by decoding $F_o^l$ to image domain. The decoder can be pre-trained for efficiency~\cite{li2017universal,huang2017arbitrary,lu2017decoder}. By contrast, we adopt an different decoder learning from theirs. The details of decoder training are discussed in \Sref{sect:decoder}.



\begin{figure}[t]
\centering
 \includegraphics[width=0.95\linewidth]{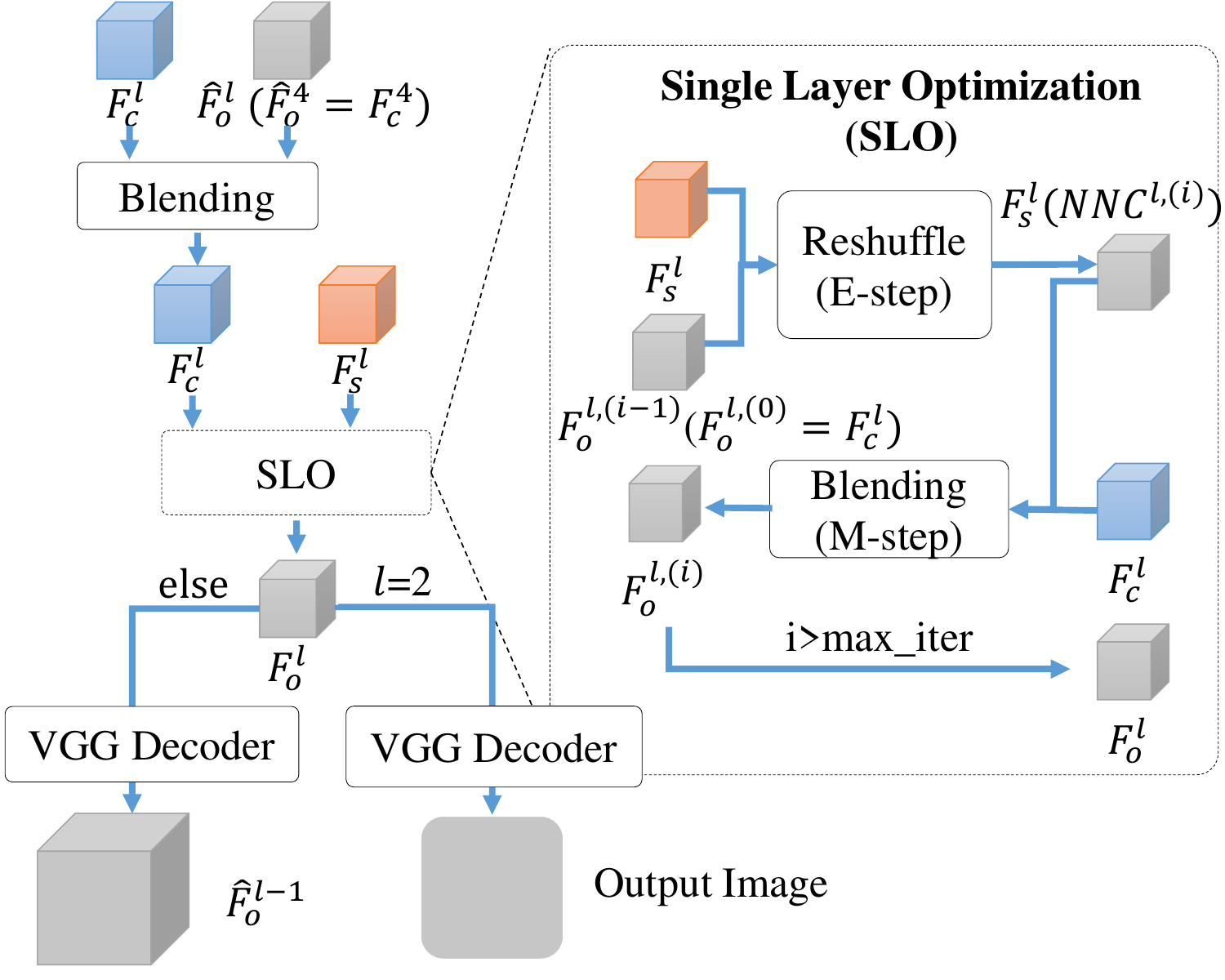}
    \caption{System Pipeline}\vspace{-0.7em}
\label{fig:method}
\end{figure}

\subsection{Multi-layer Progressive Optimization}
\label{sec:multi}
We actually adopt a multi-layer progressive optimization instead of independent single layer optimization described in the above section. This way provides three advantages. First, leveraging mutil-layer features can generate richly textured results. Second, multi-resolution processing can help avoid getting stuck in worse local minima. Last, with the NNF guidance from coarse layers, we may accordingly decrease the search range for more efficient matching.

\fref{fig:method} shows our algorithm pipeline. The multi-layer optimization considers such three layers $l=2,3,4$. In initialization, we obtain feature maps $\{F_c^l\}_{l=4,3,2}$ and $\{F_s^l\}_{l=4,3,2}$ of $I_c$ and $I_s$ respectively by feeding them to VGG-19.

We start from the coarsest layer $l=4$, and perform the single layer optimization (see~\Sref{sec:sig}) at this layer, which uses EM-like constrained patch match algorithm. After it, we get the updated feature $F_o^{4}$, which is then decoded to the next layer $l-1$ by our trained decoder network, denoted as $\hat F_o^{l-1}$. To leverage the result from the coarse layer, we update the content feature $F_c^{l-1}$ at layer $l-1$ by linearly combining it with $\hat F_o^{l-1}$, namely, $F_c^{l-1} \leftarrow \beta \hat F_o^{l-1}+(1-\beta)F_c^{l-1}$. Next, we use the updated content feature $F_c^{l-1}$ for single layer optimization in layer $l-1$. The blending can inherit information from the coarser layers. We iterate the process from the coarsest layer $l=4$ to the finest layer $l=2$. Finally, we decode the optimal feature $F_o^2$ to obtain the output image $I_o$. The pseudo code of our implementation is listed in Algorithm~\ref{ag:bda}. Code has been made available at: \url{https://github.com/msracver/Style-Feature-Reshuffle}

\begin{algorithm}[t]
\caption{The deep feature reshuffle algorithm.}
\small
\DontPrintSemicolon
\SetAlgoLined
\SetKwInOut{Input}{Input}\SetKwInOut{Output}{Output}
\Input{One content image $I_c$ and one style image $I_s$.}
\BlankLine
 \textbf{Initialization}:\\
 \quad$\{F_{c}^{l}\}_{l=2}^4, \{F_{s}^{l}\}_{l=2}^4\leftarrow$ feed $I_c, I_s$ to VGG-19.\\
 \quad$\hat F_o^4=F_c^4$.\\
\For{$l =4$ to $2$}{
   $F_c^l \leftarrow \beta \hat F_o^l+(1-\beta)F_c^l$.\\
   $F_o^{l,(0)} \leftarrow F_c^l$.\\
   \For{$i =1$ to $max\_iter$}{
   $\text{NNC}^{l, (i)} \leftarrow$ match $F_o^{l, (i-1)}$ and $F_s^l$ (\eref{eq:nnc}).\\
   $F_s^l(\text{NNC}^{l,(i)}) \leftarrow$ warp $F_{s}^{l}$ with $\text{NNC}^{l,(i)}$.\\
   $F_o^{l,(i)} \leftarrow \alpha F_c^l+(1-\alpha)F_s^l(NNC^{l,(i)})$.\\
   }
   \If{$l>2$}{
   $\hat F_o^{l-1} \leftarrow$ decode $F_o^{l,(max\_iter)}$ from $l$ to $l-1$.
   }
   }
$I_o \leftarrow$ decode $F_o^{2,(max\_iter)}$ to image.\\
\Output{Style transfer result image $I_o$. }
\label{ag:bda}
\end{algorithm}

\subsection{Decoder Training}\label{sect:decoder}

\begin{figure}[t]
\centering
 \includegraphics[width=0.95\linewidth]{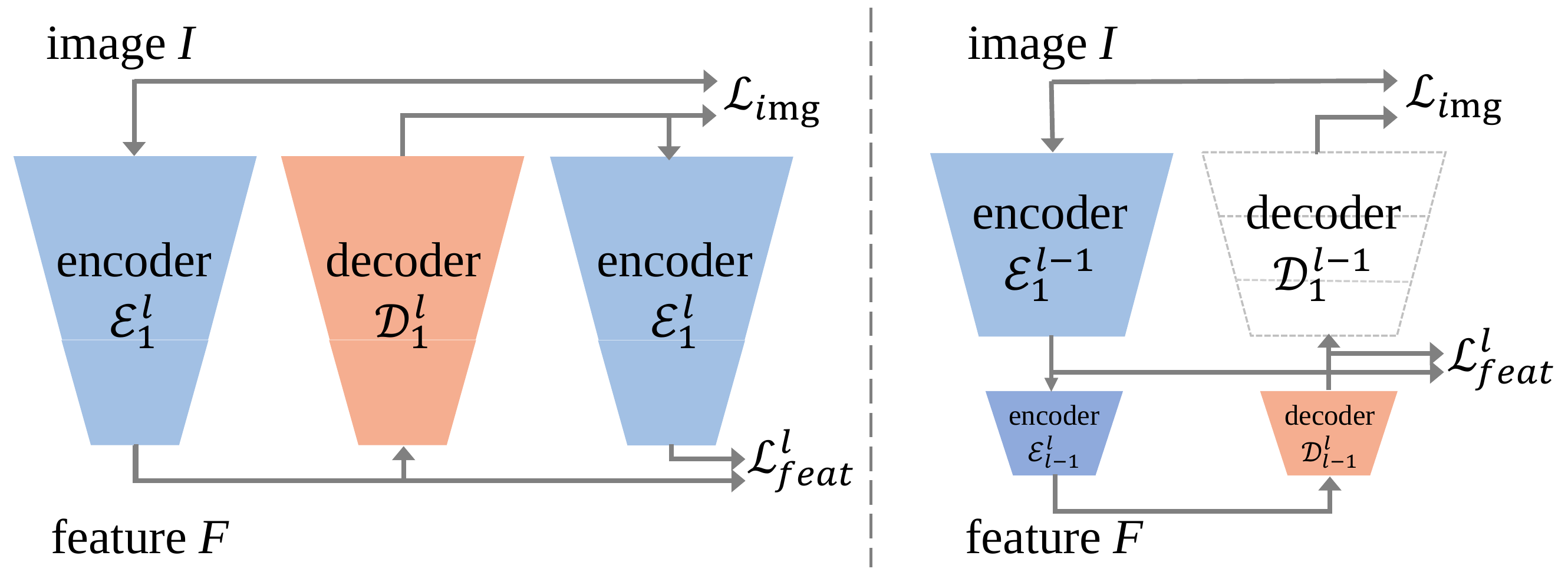}
    \caption{Comparison of our decoder (right) and Li et al.'s~\cite{li2017universal} (left). To train a decoder in layer $l$, ours fixes and shares the pre-trained part from layer $1$ to $l-1$, while theirs retrains all parts. }\vspace{-1.0em}
\label{fig:decoder}
\end{figure}

Li et al.~\cite{li2017universal} proposed a universal decoder for fast style transfer. However, their method is not very economic, since $L$ various decoders are needed to respectively decode features from every different layer. These decoders do not share weights in training. In this paper, we propose a new training strategy which provides only a single decoder for features may from different layers. The comparisons between the two training strategies are shown in ~\fref{fig:decoder}.

\begin{figure}[t]
\centering
\footnotesize
\setlength{\tabcolsep}{0.003\linewidth}
\begin{tabular}{cccc}
 \includegraphics[width=0.22\linewidth]{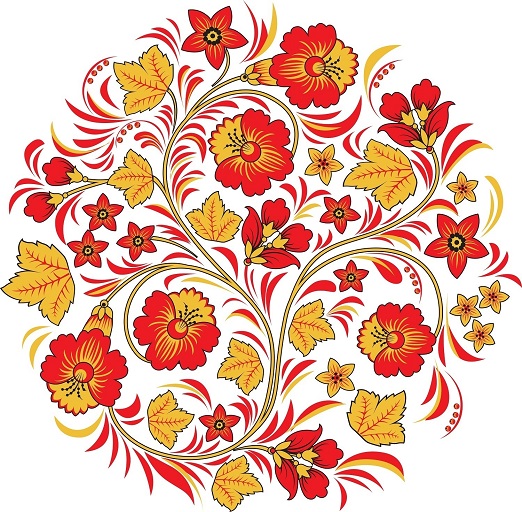} &  \includegraphics[width=0.22\linewidth]{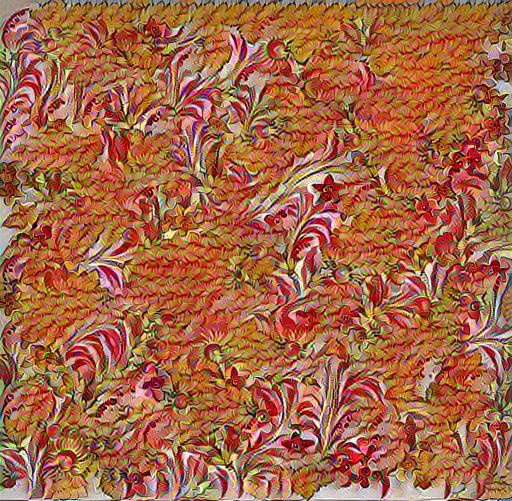} & \includegraphics[width=0.22\linewidth]{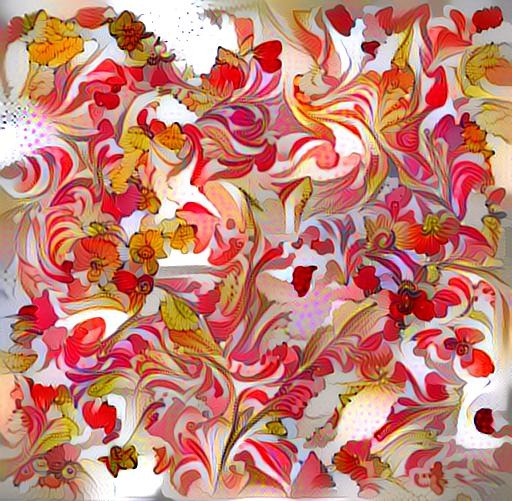} & \includegraphics[width=0.22\linewidth]{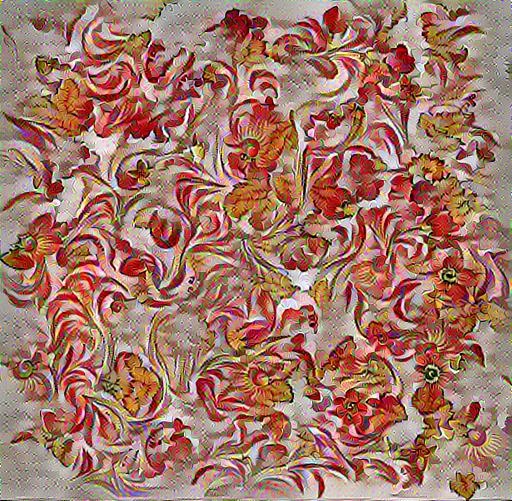}\\
 \includegraphics[width=0.22\linewidth]{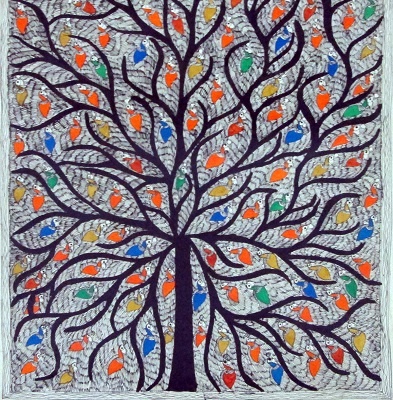} &  \includegraphics[width=0.22\linewidth]{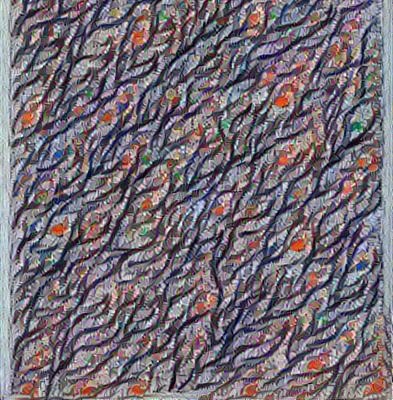} & \includegraphics[width=0.22\linewidth]{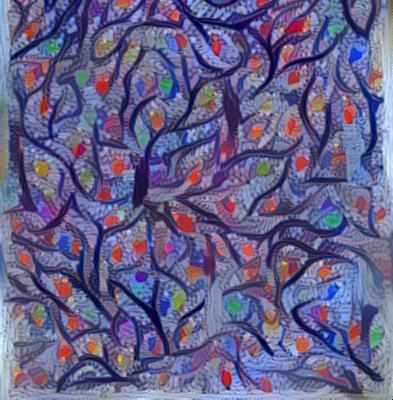} & \includegraphics[width=0.22\linewidth]{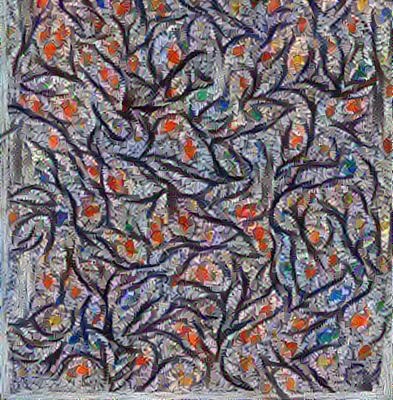}\\
 Texture & Non-parametric\cite{chuanli2016} & Parametric\cite{gatys2015neural}  & Ours  \\
\end{tabular}       \caption{Texture synthesis results by minimizing different style losses on layer $4$}
\vspace{-1.0em}
\label{fig:texture_systhesis}
\end{figure}

The architecture of our decoder uses a symmetrical structure to that of VGG-19 encoder network. The training strategy is bottom-up, starting from layer $1$ to layer $L$.  When we train the $l\_\text{th}$ layer of decoder, image $I$ first feeds to the encoder sub-net $\mathcal{E}_1^{l-1}$ involving encoder layer $1$ to $l-1$, and achieves feature map at layer $l-1$, $F^{l-1} = \mathcal{E}_1^{l-1}(I)$. It will further feed to the encoder $\mathcal{E}_{l-1}^{l}$ involving encoder layer $l$, and get feature $F^{l} = \mathcal{E}_{l-1}^{l}(F^{l-1})$. Here, our loss includes two folds. On one hand, feature $F^{l}$ would directly feed to the decoder for image reconstruction, namely $\hat I = \mathcal{D}_{1}^{l-1}(\hat F^{l-1})$, where $\hat F^{l-1} = \mathcal{D}_{l-1}^{l}(F^{l})$. We use $L_2$-norm based image reconstruction loss, $\mathcal{L}_{img} = ||\hat I - I||^2$. On the other hand, we measure feature loss by $\mathcal{L}_{feat} = ||\hat F^{l-1} - F^{l-1}||^2$. In summary, the decoder sub-net $\mathcal{D}_{l-1}^{l}$ only involving decoder layer $l$, can be achieved by $\min (\mathcal{L}_{img} + \mathcal{L}_{feat})$, which can be rewritten as:
$\min(||I-\mathcal{D}_{1}^{l-1}(\mathcal{D}_{l-1}^{l}(F^{l}))||^2 + ||F^{l-1} - \mathcal{D}_{l-1}^{l}(F^{l}) ||^2),
$
where the $\mathcal{D}_{1}^{l-1}$ is fixed when $\mathcal{D}_{l-1}^{l}$ is computed in our training. By contrast, both $\mathcal{D}_{1}^{l-1}$ and $\mathcal{D}_{1-1}^{l}$ are always retrained for every decoder layer~\cite{li2017universal}. We train our decoder on the ImageNet dataset \cite{krizhevsky2012imagenet}.

\section{Ablation Study}
\label{sec:abla}

\subsection{Style Loss Analysis}

We study the effect of different style loss terms, including global style loss (in~\eref{eq:global}), local style loss (in~\eref{eq:local}), and reshuffle style loss (in~\eref{eq:shuf}), by neglecting the common content loss. Thus, we only evaluate them on texture synthesis. We collect 60 image pairs from existing papers, and start from random noise to respectively minimize the three losses in layer $4$. The optimization is conducted by L-BFGS method, and stopped at 500 iterations, where results have no visible changes with further iterations.

\begin{figure}[t]
\centering
 \includegraphics[width=0.95\linewidth]{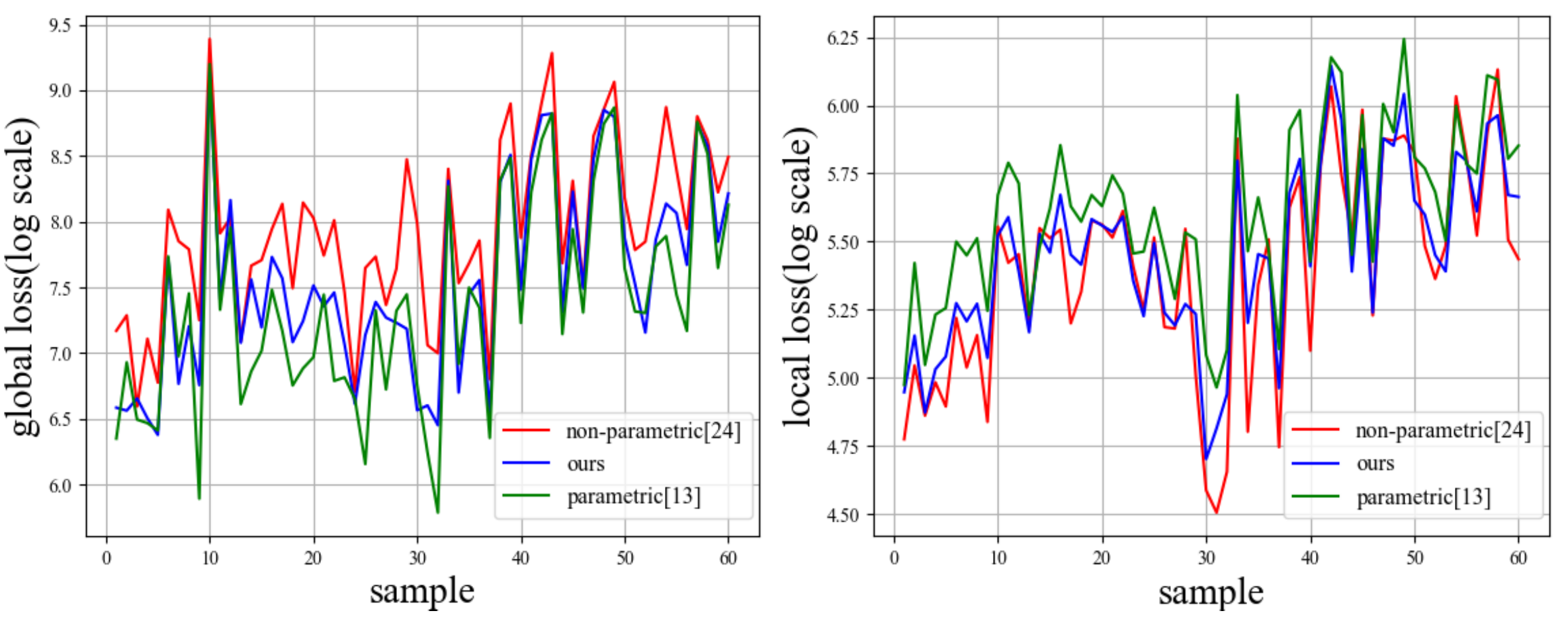}
    \caption{Global style loss (\eref{eq:global}) (left) and local style loss (\eref{eq:global}) (right) measured on every result with various methods. }\vspace{-1.0em}
\label{fig:global_local}
\end{figure}

Some results are shown in~\fref{fig:texture_systhesis}. As we can see, the results by minimizing the global style loss (\eg,~\cite{gatys2015texture}) better reproduce the overall feeling of the style, while the results by minimizing the local style loss (\eg,~\cite{chuanli2016}) are more faithful to the local shapes of example texture. The results using our reshuffle loss own both merits: global similarity and local plausibility. The quantitative comparison also demonstrate the same point, as shown in~\fref{fig:global_local}. We can see that our loss function achieves lower global loss (\eref{eq:global}) than non-parametric method; while obtains lower local loss (\eref{eq:local}) than parametric method, in all the cases.

\begin{figure}[t]
\centering
 \includegraphics[width=0.95\linewidth]{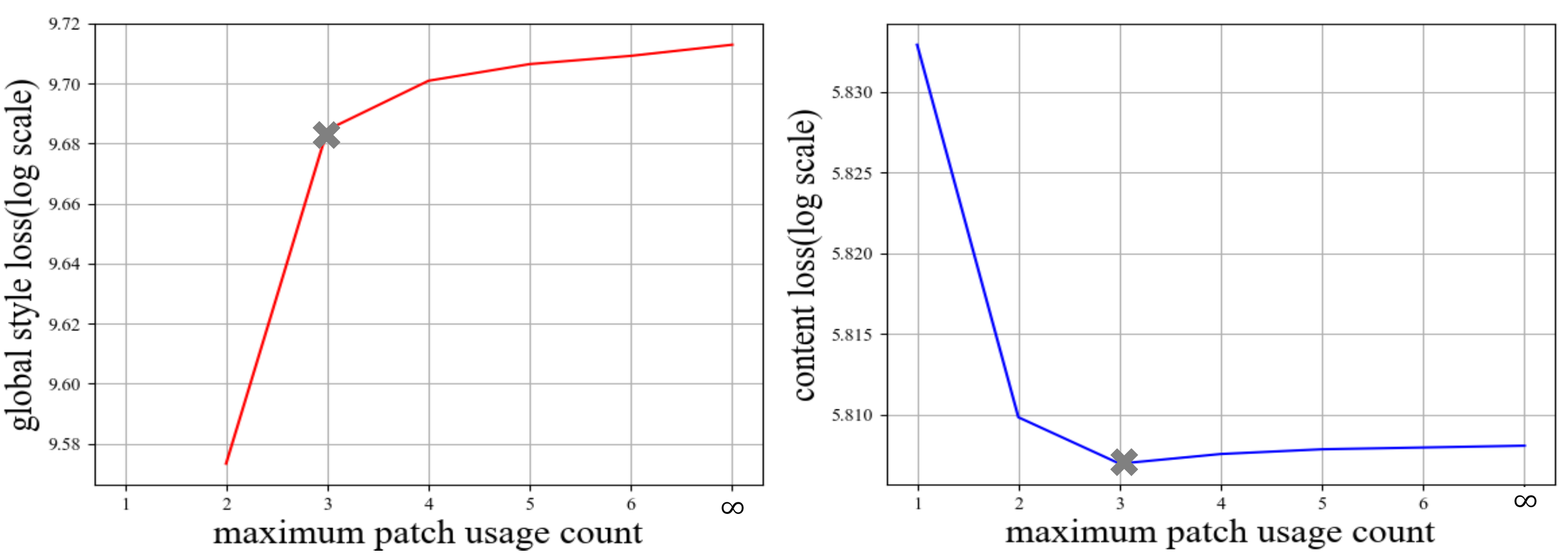}
    \caption{The global style loss (left) increases while content loss (right) decreases with the increase of maximum patch usage count.} \vspace{-0.5em}
\label{fig:usage}
\end{figure}

\subsection{Count of Patch Usage}
\label{sec:use}

\begin{figure}[t]
\centering
\footnotesize
\setlength{\tabcolsep}{0.005\linewidth}
\begin{tabular}{ccc}
 \includegraphics[width=0.3\linewidth]{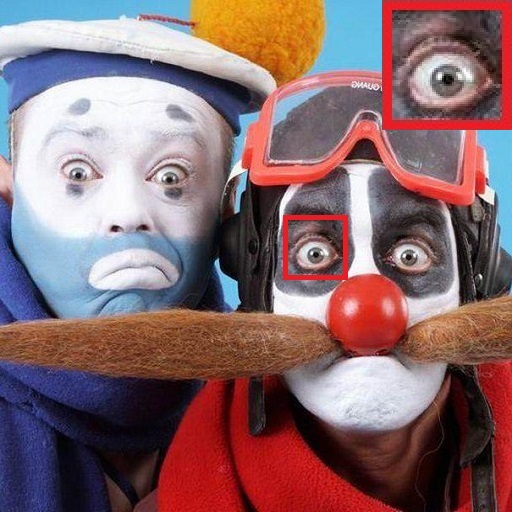} &  \includegraphics[width=0.3\linewidth]{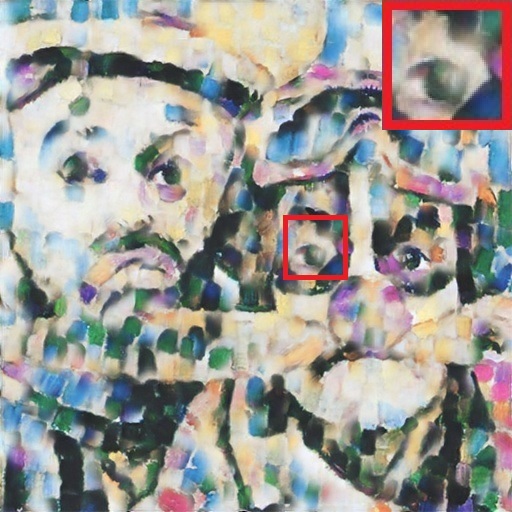} & \includegraphics[width=0.3\linewidth]{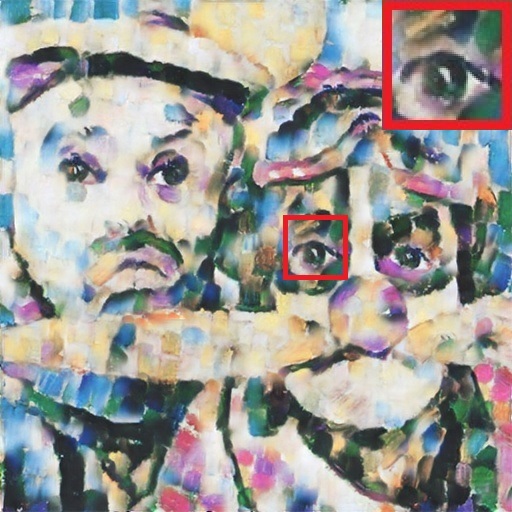}\\
 Content& Max usage $= 1$ & Max usage $= 2$\\
\includegraphics[width=0.3\linewidth]{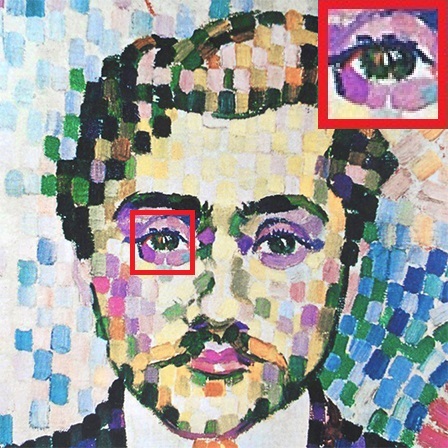} &  \includegraphics[width=0.3\linewidth]{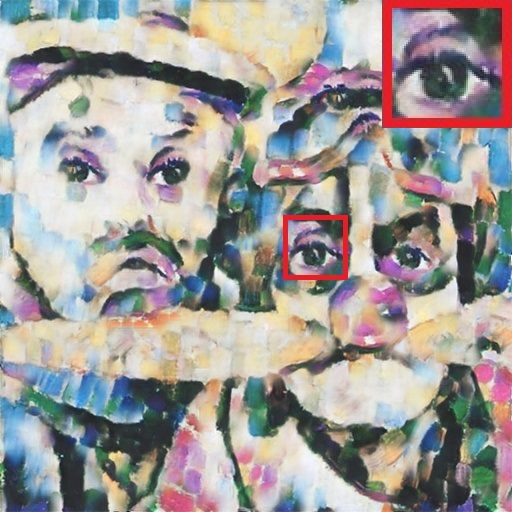} & \includegraphics[width=0.3\linewidth]{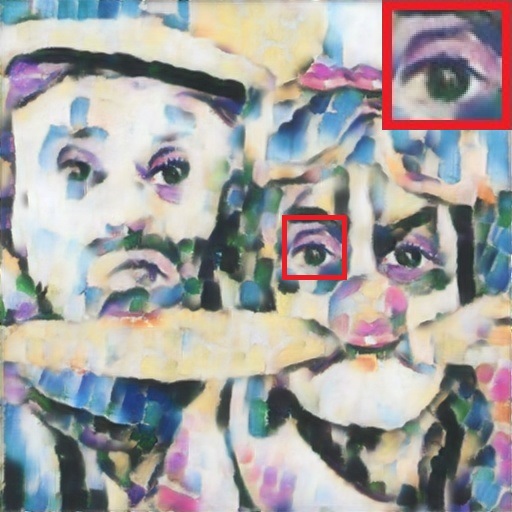}\\
Style & Max usage $= 3$ & Soft ($\lambda=0.05$)\\
\end{tabular}
    \caption{A comparison of results with different usage count constraints. Please note the eye in the red rectangles are miss-matched when max usage $= 1$, but fixed when max usage increases. }\vspace{-0.1in}
\label{fig:usage2}

\end{figure}

We will examine the effect of patch usage count. The relaxation of maximum usage count only reduces global style loss, but the local style loss always remains 0. We still consider the above 60 examples, and use our single layer optimization (in~\Sref{sec:sig}, only layer $4$) to compute the content loss (\eref{eq:content}) and the global style loss (\eref{eq:global}). Here, we try varied maximum counts of patch usage, and shows their corresponding content losses and global style losses in \fref{fig:usage}. With the increase of maximum usage count, the global style loss increases while the content loss decreases. It is not hard to understand that allowing more usage times will provide more choices to match content. Moreover, we find the good upper bound for patch usage is $3$, with minimum style loss given the best preservation to content, which helps infer a tradeoff weight $\lambda=0.05$ in \eref{eq:nnc}. A visual comparison of different usage constraints is shown in \fref{fig:usage2}.

 \begin{figure}[t]
\centering
 \includegraphics[width=0.92\linewidth]{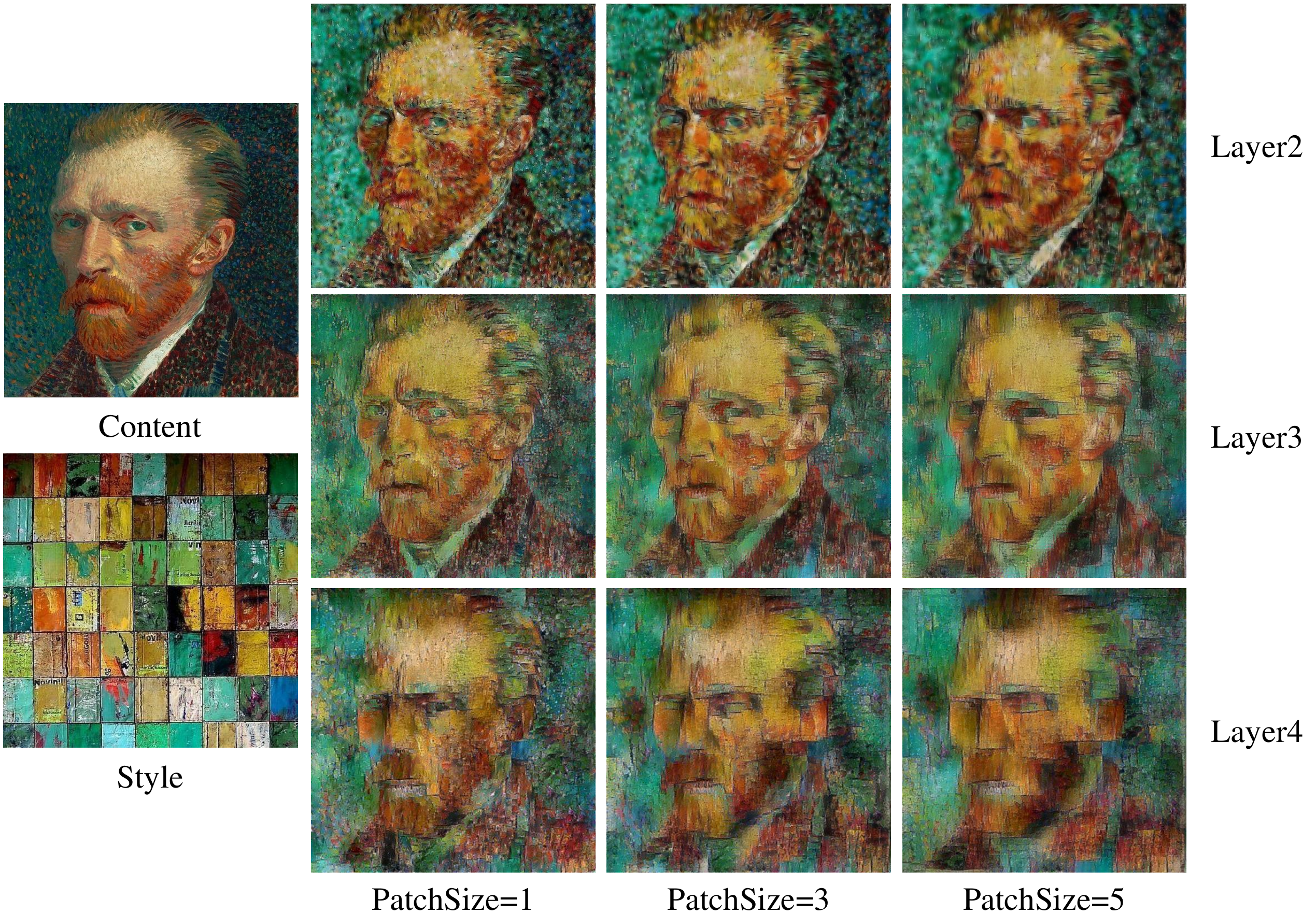}
    \caption{An example of style transfer results with different patch sizes and at different layers.}\vspace{-0.1in}
\label{fig:patchsize}
\end{figure}

\subsection{Patch Size Selection}

Another hyperparameter in our method is the patch size $R$. Increasing patch-size will sacrifice the global style loss (in~\eref{eq:global}) to some extend. However, in some scenarios, large patch size is needed to preserve spatial coherence. As shown in \fref{fig:patchsize}, lager patch size we use, better local structure of style patterns can be preserved. Another factor is that patches in coarser layers may cover larger reception fields, making good matching difficult. So we choose empirically $3\times3$ patch in layer $4$ and $5\times5$ patch in layers $2, 3$.

\section{Results}
 
\subsection{Implementation Details}
\label{sec:imp}
All results are produced by multi-layer aggressive optimization in feature domain (in \Sref{sec:multi}). On each layer, we do 5 EM iterations. The used patch sizes are $\{R^l\}_{l=2,3,4}=\{5,5,3\}$ and the patch usage parameter is set to be $\lambda = 0.05$ respectively according to the experiment in \Sref{sec:abla}. As to the weight for content and style balancing, we set $\alpha=0.5$ and $\beta=1.0$ to make our stylization level similar to previous works.

\begin{figure*}[t]
\centering
\footnotesize
\includegraphics[width=0.95\linewidth]{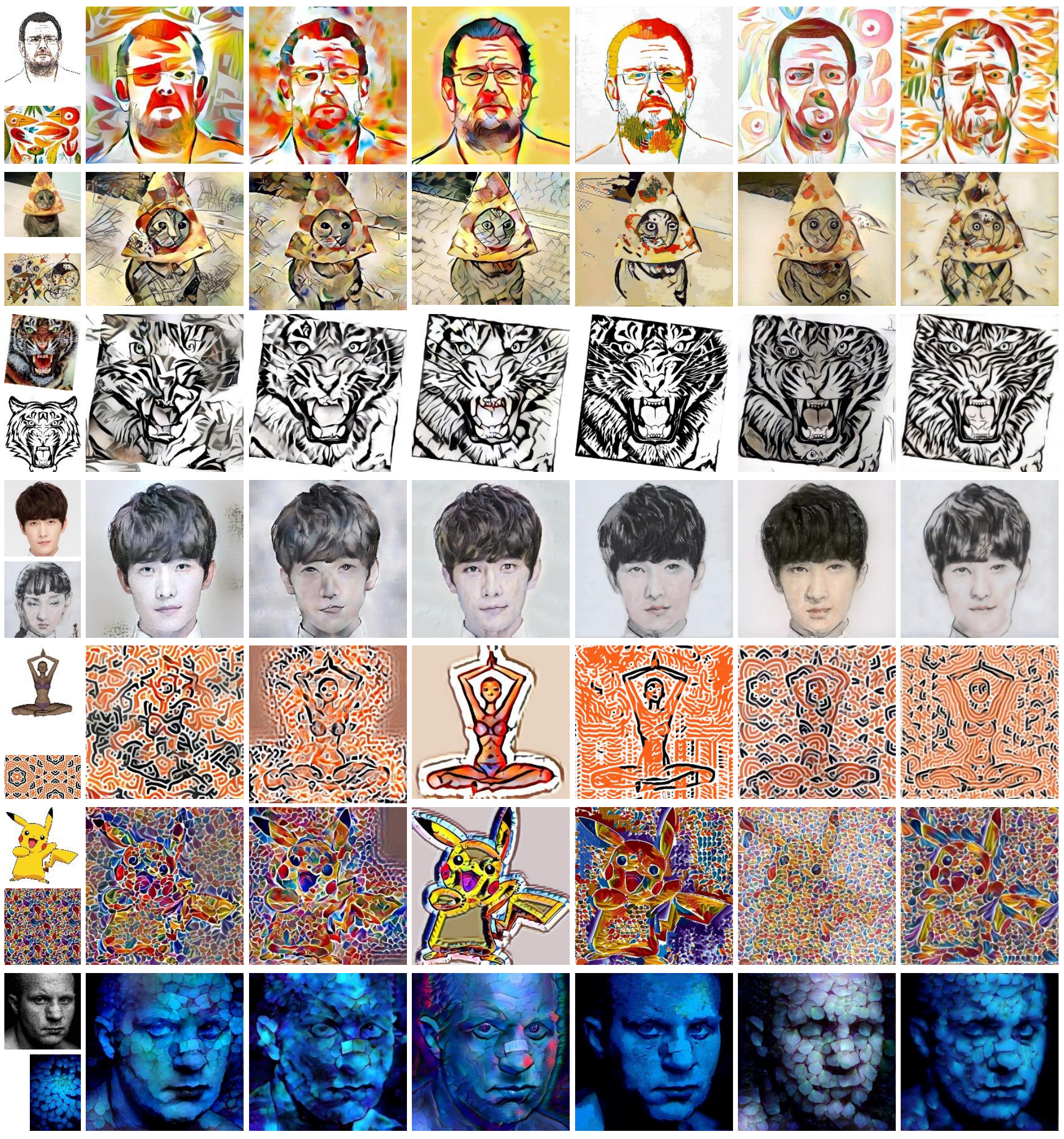}
\begin{tabular}{l|c|c|c|c|c|c}
Method&\ \ \ Gatys et al.\cite{gatys2015neural}\ \ \ &\ \ \ \ \ \  Li et al.\cite{li2017universal} \ \ \ \ \  &\ \ Huang et al.\cite{huang2017arbitrary} \ \ &\ \ \ \ Liao et al.\cite{liao2017visual}\ \ \ \ \ &\ \ \ \ \ \ Li et al.\cite{chuanli2016}\ \ \ \ \ \  & \ \ \ \ \ \ \ \ \ \ \ \ Ours\ \ \ \ \ \ \ \ \ \ \ \ \\
\hline
Time(s)&370&8.46&0.556&114&195&114\\
\end{tabular}
\caption{Comparison with previous neural style transfer methods.}
\label{fig:resultfig}
\end{figure*}

\subsection{Comparisons}
We compare our result with other neural style transfer methods including representative parametric methods \cite{gatys2015neural,huang2017arbitrary,li2017universal} and non-parametric methods \cite{chuanli2016,liao2017visual}. For fair comparison, all our results are generated with fixed parameters as described in \Sref{sec:imp}. And theirs are obtained by running author-released code with default settings. We have tested on more than 100 content and style pairs collected from previous papers and Ostagram website \cite{Ostgram}. \fref{fig:resultfig} shows some representative results. More results can be found in our supplemental material.

As shown in \fref{fig:resultfig}, our method shares advantages from both kinds of methods. First, as a non-parametric method, our results preserve the local texture patterns better, more faithful to the style, compared with parametric methods~\cite{gatys2015neural,huang2017arbitrary,li2017universal}, as shown in row $1\&2$ of \fref{fig:resultfig}. Unfortunately, in their results, the local textures are distorted, and some new patterns (not belongs to the style) appears. Second, our method seeks for best matches for each local patch in the content, so it can better achieve semantic-level transfer (\eg, eye-to-eye, mouth-to-mouth) than parametric methods which only mimic the global statistics \cite{gatys2015neural,huang2017arbitrary,li2017universal}, as shown in row $3\&4$ of \fref{fig:resultfig}. Third, our method can own good global properties, making it different from other non-parametric methods~\cite{chuanli2016,liao2017visual}. On one hand, our result can better preserve overall feels of the exemplar style, as shown in the row $5\&6$ of \fref{fig:resultfig}. Non-balanced neural patch sampling~\cite{chuanli2016,liao2017visual} makes their results  different from the global distribution of the style patterns; while ours are globally more faithful. On the other hand, our method can successfully avoid excessively repetitive use of the same sample, which will cause the washout effect \cite{jamrivska2015lazyfluids}. We can clearly see these undesired effects in row $2\&7$ of Liao et al.'s results \cite{liao2017visual}. The property benefits from our reshuffle constraint.

We also compare the time cost of all these methods. Tabel below \fref{fig:resultfig} gives the average running time of each method on $512\times 512$ image pairs. All the methods are tested on a PC with an Intel E5 2.6GHz CPU and an NVIDIA Tesla K40C GPU. Our method is slower than \cite{huang2017arbitrary,li2017universal}, comparable to \cite{liao2017visual} and faster than \cite{gatys2015neural,chuanli2016}. The bottleneck is the constrained NN field search step.

\subsection{Perceptual Study}
We conduct a user study similar to~\cite{jing2017neural}. In the study, we use 150 groups of images shown in the supplemental material. Each group contains two inputs and six outputs (involving 5 results from \cite{gatys2015neural,li2017universal,huang2017arbitrary,liao2017visual,chuanli2016}, and ours). All six results in each group are presented side-by-side and in a random order to participants. Participants are given unlimited time to rank the score from 1 to 6 (1 is the best, 6 is the worst) according to preference. We show the average ranking scores over 15 participants in Table \ref{table:perceptual-study}. Overall, subjects prefer our result more than others.

\begin{table}[h]
\vspace{-1pt}
\centering
\caption{Average stylization rank scores of six algorithms}\label{table:perceptual-study}
\small
\begin{tabular}{ccccccc}
 \hline
Method &\cite{gatys2015neural} &\cite{li2017universal} & \cite{huang2017arbitrary} &\cite{liao2017visual} &\cite{chuanli2016} & Ours\\
 \hline
Average rank& 3.08 & 3.55 & 3.92 & 3.42 & 4.2 & 2.88 \\
 \hline
\end{tabular}
\vspace{-5pt}
\end{table}

\begin{figure}[t]
\footnotesize
\setlength{\tabcolsep}{0.003\linewidth}
\centering
\begin{tabular}{cccc}
 \includegraphics[width=0.22\linewidth]{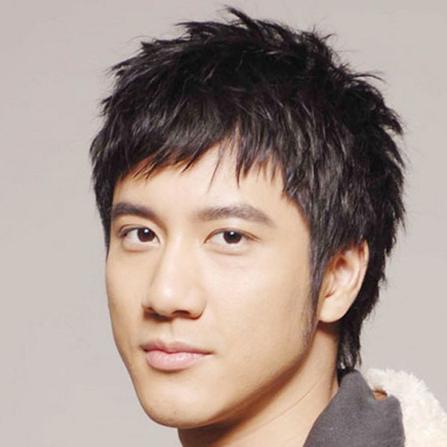} &  \includegraphics[width=0.22\linewidth]{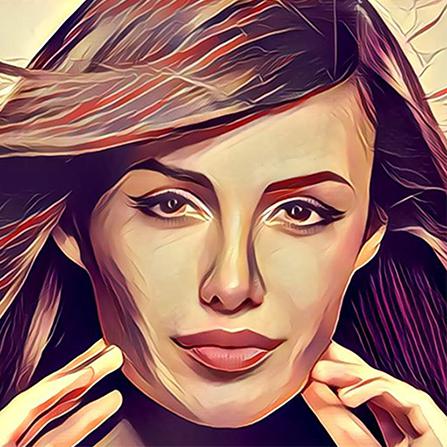} & \includegraphics[width=0.22\linewidth]{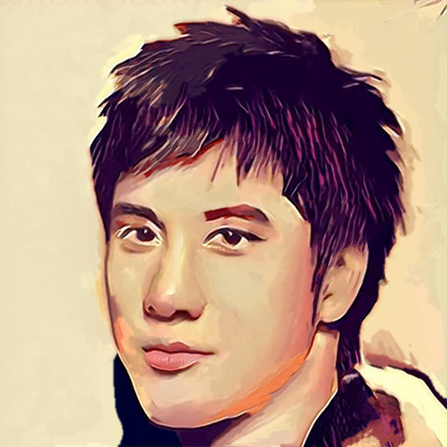} & \includegraphics[width=0.22\linewidth]{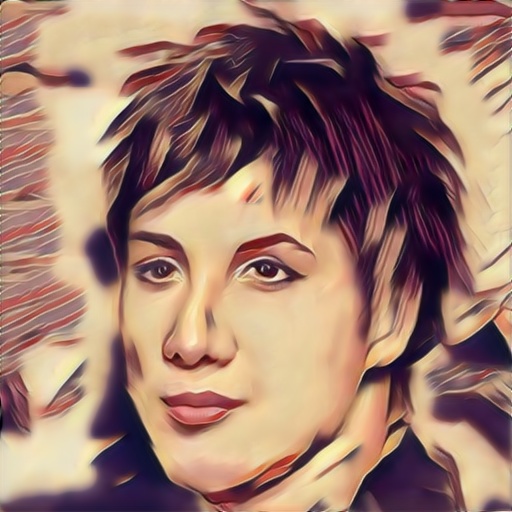}\\
 Content & Style & Liao et al. \cite{liao2017visual} & Ours\\
 \end{tabular}
 \caption{A failure case.}\vspace{-0.15in}
 \label{fig:limit}
 \end{figure}
 
\section{Discussion and Conclusion}
Despite the success of neural style transfer, the relationship between different methods was far from clear. In this paper, we give a new perspective to connect them. We then propose a novel and simple idea, called deep feature reshuffle, which is the first to unify both commonly-used global and local style losses. Based on this idea, we propose a new and efficient neural style transfer algorithm by progressively optimizing the new loss in feature domain. The results have shown that our approach is widely applicable to various inputs, and produces better quality than existing methods.

However, the method still suffers from some limits. Constraining the usages of neural patch for the sake of style, will cause less accurate matching and thus damage the content structure, as shown in \fref{fig:limit}. It can be solved by fine-tuning the usage parameter $\lambda$. How to automatically determine the optimal parameter for each input will become a vital and practical problem to be explored in future work.

{\small
\bibliographystyle{ieee}
\bibliography{egbib}
}

\end{document}